\documentclass[sigplan,screen]{acmart}

\usepackage{subcaption} 
\usepackage{multirow}

\AtBeginDocument{%
  }

\begin{document}

\title{Towards Training-Free Open-World Classification with 3D Generative Models}

\author{Xinzhe Xia}
\authornote{Both authors contributed equally to this research.}
\affiliation{%
  \institution{Xi’an Jiaotong-Liverpool University}
  \city{Suzhou}
  \state{Jiangsu}
  \country{China}
}
\affiliation{%
  \institution{University of Liverpool}
  \city{Liverpool}
  \state{}
  \country{UK}
}
\email{Xinzhe.Xia23@student.xjtlu.edu.cn}

\author{Weiguang Zhao}
\authornotemark[1]
\affiliation{%
  \institution{Xi’an Jiaotong-Liverpool University}
  \city{Suzhou}
  \state{Jiangsu}
  \country{China}
}
\affiliation{%
  \institution{University of Liverpool}
  \city{Liverpool}
  \state{}
  \country{UK}
}
\email{weiguang.zhao@liverpool.ac.uk}

\author{Yuyao Yan}
\authornote{Corresponding author}
\affiliation{%
  \institution{Xi’an Jiaotong-Liverpool University}
  \city{Suzhou}
  \state{Jiangsu}
  \country{China}
}
\email{Yuyao.Yan@xjtlu.edu.cn}

\author{Guanyu Yang}
\affiliation{%
  \institution{Xi’an Jiaotong-Liverpool University}
  \city{Suzhou}
  \state{Jiangsu}
  \country{China}
}
\email{Guanyu.Yang02@xjtlu.edu.cn}

\author{Rui Zhang}
\affiliation{%
  \institution{Xi’an Jiaotong-Liverpool University}
  \city{Suzhou}
  \state{Jiangsu}
  \country{China}
}
\email{Rui.Zhang02@xjtlu.edu.cn}

\author{Kaizhu Huang}
\affiliation{%
  \institution{Duke Kunshan University}
  \city{Kunshan}
  \state{Jiangsu}
  \country{China}}
\email{kaizhu.huang@dukekunshan.edu.cn}

\author{Xi Yang}
\affiliation{%
  \institution{Xi’an Jiaotong-Liverpool University}
  \city{Suzhou}
  \state{Jiangsu}
  \country{China}
}
\email{xi.yang01@xjtlu.edu.cn}

% \author{
%     Xinzhe Xia$^{1,2}$\thanks{Equal contribution.}\and
%     Weiguang Zhao$^{1,2}$\footnotemark[1]\and
%     Yuyao Yan$^{1}$\thanks{Corresponding author.}\and
%     Guanyu Yang$^{1}$\and
%     Rui Zhang$^{1}$\and \\
%     Kaizhu Huang$^{3}$\and 
%     Xi Yang$^{1}$\\
%     \affiliations
%     $^1$Xi'an Jiaotong-Liverpool University, $^2$University of Liverpool, $^3$Duke Kunshan University
%     \emails
%     \{Xinzhe.Xia23, Weiguang.Zhao20\}@student.xjtlu.edu.cn \\
%     \{Yuyao.Yan, Guanyu.Yang02, Rui.Zhang02, Xi.Yang01\}@xjtlu.edu.cn
% }

\renewcommand{\shortauthors}{Trovato et al.}

\begin{abstract}
3D open-world classification is a challenging yet essential task in dynamic and unstructured real-world scenarios, requiring robust subsequent knowledge adaptation capabilities. While current approaches predominantly rely on 2D pre-trained models through 3D-to-2D projection, their performance degrades severely under arbitrary object orientations. 
Unlike these present efforts, this work makes a pioneering exploration of 3D generative models for 3D open-world classification-specifically, leverageing the accumulated prior knowledge from these models to provide anchors for novel categories, while integrating a rotation-invariant feature extractor. 
This innovative synergy endows our pipeline with the advantages of being training-free and pose-invariant, thus well suited to adapt novel categories in 3D open-world classification.
Extensive experiments on benchmark datasets demonstrate the potential of this pipeline, achieving state-of-the-art performance on ModelNet10$^{\ddagger}$ and McGill$^{\ddagger}$ with 32.7\% and 8.7\% overall accuracy improvement, respectively. The code is available in the supplementary materials. 
\end{abstract}




% \received{20 February 2007}
% \received[revised]{12 March 2009}
% \received[accepted]{5 June 2009}

\maketitle

\section{Introduction}
%Research into 
3D classification in open-world environments presents critical challenges in developing computational models that can accurately interpret complex, unstructured scenarios while adapting to novel knowledge. Such advances are crucial for improving the autonomy and adaptability of intelligent systems~\cite{NEURIPS2023_8c7304e7,zhang2023clip,zhao2024towards}.
This field aims to effectively categorize objects that not only vary in trained and novel categories--a challenge of open-category recognition--but also appear in arbitrary poses, posing an open-pose recognition dilemma~\cite{zhang2023clip,Zhu2022PointCLIPV2,open-pose}. 

Knowledge adaptation in current open-world 3D classification approaches is based mainly on three paradigms: language-only, language-point, and language-image model-based strategies~\cite{open-pose}.  While promising, each suffers from limitations that motivate our investigation. Language-only model-based methods~\cite{cheraghian2020transductive,cheraghian2022zero} rely solely on textual information to infer connections with novel classes, often fall short due to \textit{the limited descriptive power of text} and \textit{the oversight of essential spatial information} inherent in 3D data.
Language-point model-based methods~\cite{xue2023ulip,xue2024ulip,xu2025pointllm}, on the other hand, \textit{demands extensive data and significant computational resources for training}, posing practical limitations. Language-image model-based methods~\cite{Zhu2022PointCLIPV2,huang2023clip2point,CLIPgoes3D,diffclip,open-pose} further complicate matters by \textit{requiring intricate projection hyperparameters} to effectively map 3D data to 2D images. In particular, all existing approaches exhibit sensitivity to pose changes.

\begin{figure}[t] 
  \centering
  \includegraphics[width=0.49\textwidth]{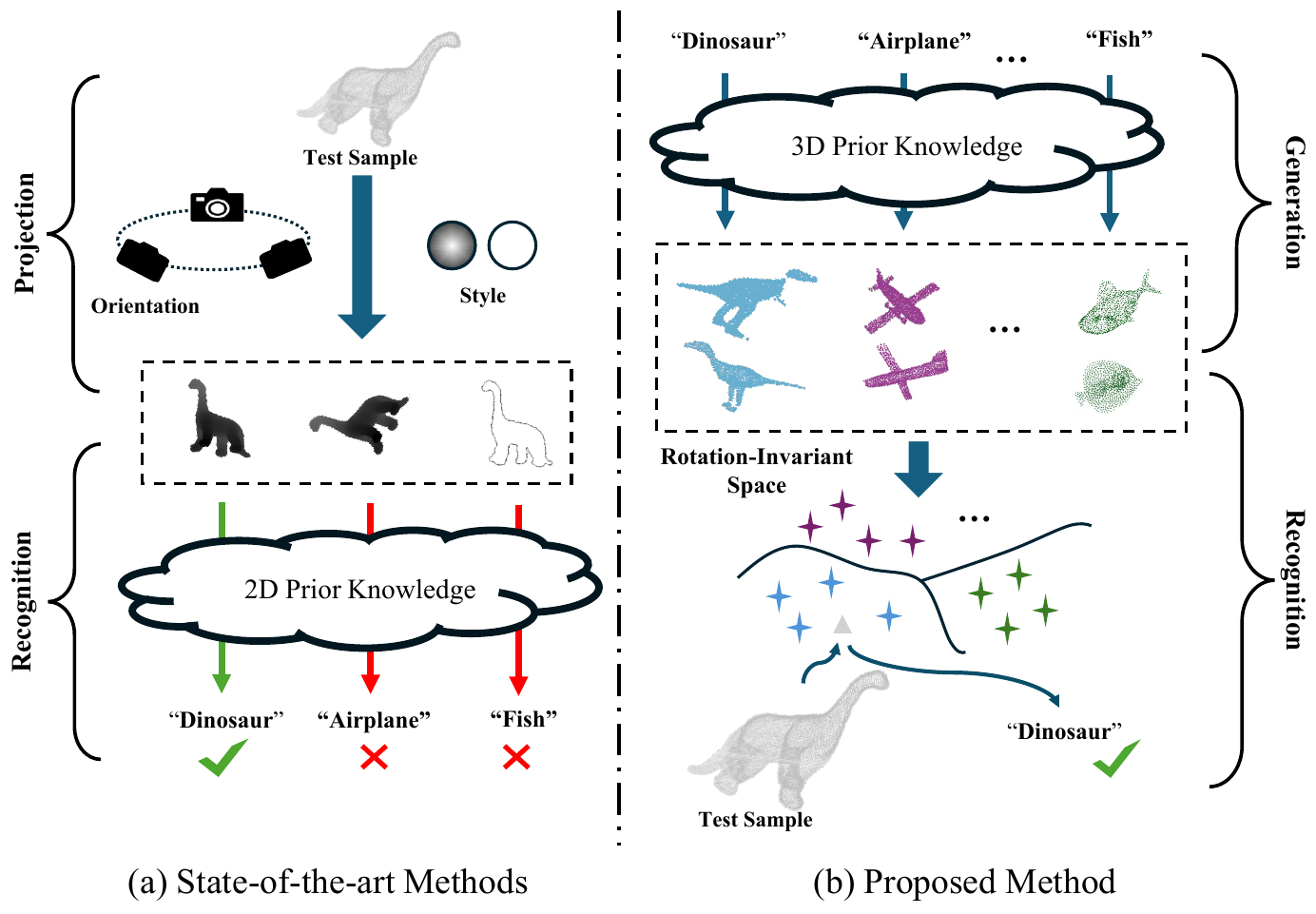} 
  \caption{Comparison of Pipelines: (a) SOTA methods~\cite{open-pose,Zhu2022PointCLIPV2} project 3D samples into 2D images to harness 2D prior knowledge for novel category recognition but are sensitive to pose changes, (b) The proposed pipeline applies 3D prior knowledge to generate anchor samples for novel categories, embedding them in a rotation-invariant space for effective performance in open-pose scenarios. Predictions are computed via feature similarity between test and anchor samples.
} 
  \label{fig:banner}
\end{figure}

Addressing these limitations requires meeting two critical demands: 1) effectively exploiting the spatial information in 3D data while maintaining computational efficiency, and 2) achieving strong adaptation capability for novel knowledge in dynamic environments. Current approaches typically require extensive retraining when encountering new scenarios - a significant drawback in open-world settings where environmental factors (like object orientation) constantly evolve.

In response to these challenges, we present a novel training-free framework that synergistically leverages pre-trained 3D generative models~\cite{db1,db2,db3,db4} with representation learning to construct a lightweight yet efficient 3D classifier for open-world scenarios.  Our proposed pipeline capitalizes on 3D prior knowledge to generate anchor samples for novel categories, subsequently embedding them into a rotation-invariant space.
As shown in Fig.~\ref{fig:banner}, unlike the state-of-the-art (SOTA) methods that rely on projecting 3D data into 2D space to exploit limited 2D priors from vision-language models, our method mitigates two critical limitations: 1) the inherent information loss from 3D to 2D projection that particularly handicaps pose-variant recognition, and 2) the computational overhead of continuous model retraining for new scenarios.

The execution of our pipeline begins with feeding textual descriptions into a pre-trained text-to-3D generation model, which synthesizes anchor samples serving as prototypes for each category. Subsequently, a pre-trained point cloud representation model is utilized to extract anchor features from these samples. These anchor features are indicative of the classification capacity of our developed classifier, as its performance exhibits a positive correlation with the quality and diversity of these generated samples (see Section ~\ref{chap:result_analysis}). Ideally, a sufficiently generalized generator will enhance the proficiency of our classifier in accurately categorizing any 3D objects. During the inference phase, cosine similarity within the representation space is computed to evaluate the similarity between test samples and anchor samples, thus leading to category predictions for the test samples. In addition, we investigate the impact of various representation models on open-world classification. Our findings suggest that most representation models fail to provide consistent representations for identical 3D objects across different orientations, displaying rotational variance. Extensive experiments on the McGill and ModelNet10 datasets validate the robustness of our approach. Additionally, we explore the use of large language models (LLMs) to generate detailed textual descriptions of category names, with the aim of enhancing limited textual descriptive capabilities.

Our contributions can be summarized as follows:
\begin{itemize}
 \item We propose a training-free open-world pipeline for 3D classification. To our best knowledge, this is the first attempt to utilize a 3D generative model as prior knowledge in the open-world setting for such tasks.
 \item We investigate the impact of the representation space within our pipeline on open-pose tasks and discuss the importance of rotation-invariant properties.
 \item Utilizing 3D point cloud augmentation techniques, we achieve state-of-the-art performance on ModelNet10$^{\ddagger}$ and McGill$^{\ddagger}$ with overall accuracy improvement of 32.7\% and 8.7\%, respectively.
\end{itemize}

\begin{figure*}[ht]  % [h] 表示尽可能在当前位置插入图片
  \centering
  \includegraphics[width=1\textwidth]{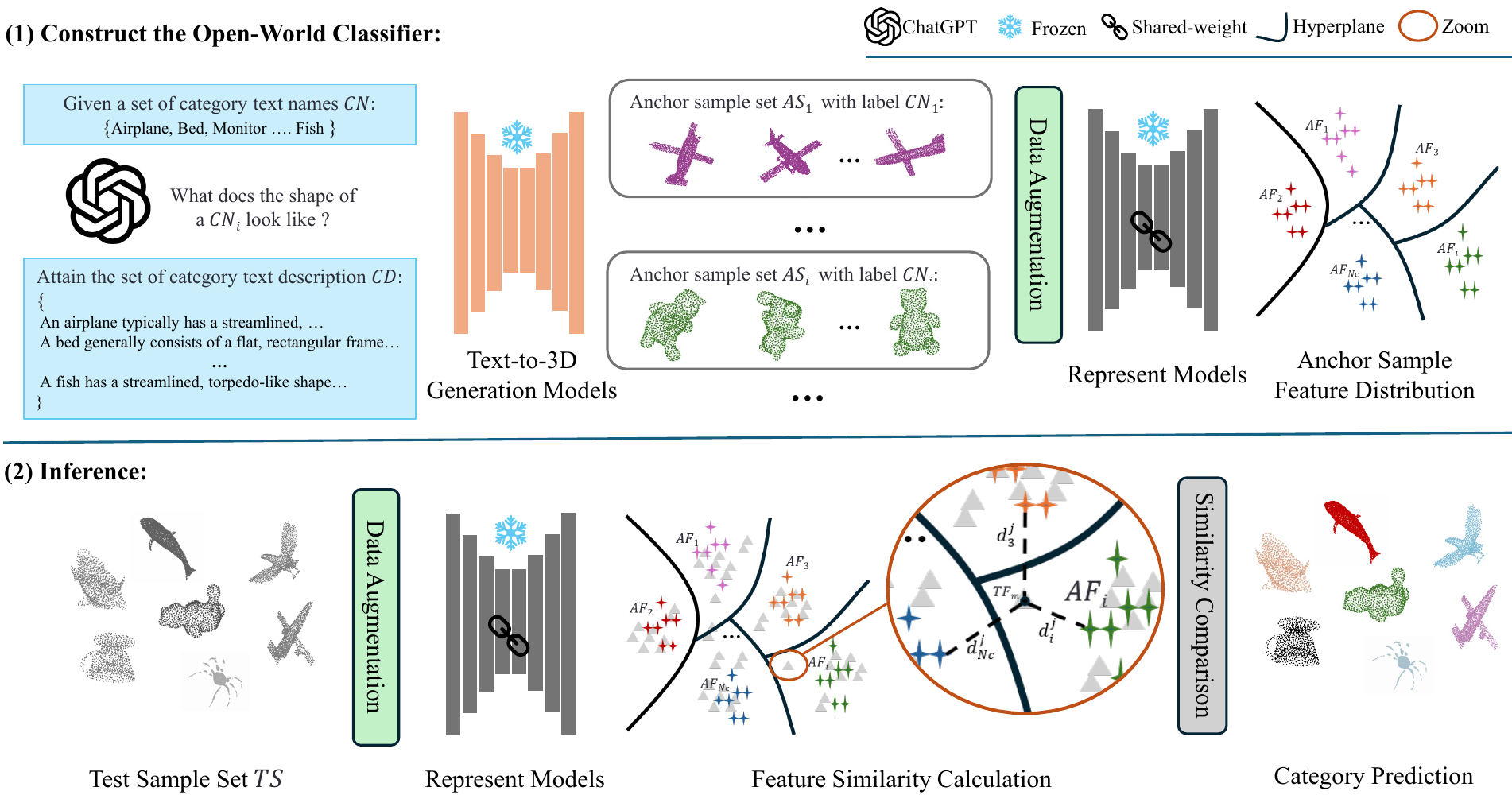}  % 插入图片并设置宽度为文本的80%
  \caption{Overview of the network architecture with two components. (1) Open-world classifier: Category names are input into ChatGPT to create descriptions, utilized by a pre-trained text-to-3D model to generate anchor samples. These samples are augmented to align with test samples. A pre-trained 3D model extracts features, marking them as prototypes for classification. (2) Test sample inference: Test samples are augmented and processed through the same model for feature extraction. Predictions are based on cosine similarity between test and anchor sample features. Different colors denote new categories, gray represents test data.}  % 图片的标题
  \label{fig:network}  % 设置图片标签，方便引用
\end{figure*}

\section{Related Work}
\subsection{3D Open-World Classification}
Current 3D open-world classification methods can be divided into three categories~\cite{open-pose}. \textbf{Language-only model-based methods}~\cite{cheraghian2020transductive,cheraghian2022zero,Hao2023ContrastiveGN} utilize pre-trained language models like Word2Vec~\cite{mikolov2013distributed} or GloVe~\cite{pennington2014glove} to establish semantic connections between training and novel classes but have fallen out of favor due to reliance on a single source of prior knowledge. 
\textbf{Language-image model-based methods}~\cite{Zhu2022PointCLIPV2,huang2023clip2point,CLIPgoes3D,diffclip,open-pose} project point clouds into multiple 2D views and utilize pre-trained models like CLIP~\cite{CLIP} or Diffusion~\cite{rombach2022high,sohl2015deep} for classifying these projections, enabling 3D open-world classification.  Although effective, they involve numerous hyperparameters, such as projection angles and viewpoint weights, which complicate the process and diminish robustness.  \textbf{Language-point model-based methods}~\cite{xue2023ulip,xue2024ulip,xu2025pointllm} link point clouds with text through various datasets and large language models. 
These methods need significant data resources and GPU memory, often with 7 or 13 billion parameter models. A recent study~\cite{open-pose} highlights the limitations of existing methods in handling open-pose 3D objects in real-world applications. Conversely, our research initiates an innovative approach within the framework of language-point model methodologies by investigating current text-to-3D generative models~\cite{db2,db3,db4} to address 3D open-world classification. Benefiting from the prior knowledge inherent in 3D generative models, our method achieves SOTA performance in open-pose scenarios without extensive hyperparameter tuning or large datasets for training. 

\subsection{Text-to-3D Generation}

With the success of diffusion models in image generation~\cite{db8}, studies~\cite{db1,db2,db3,db4} now integrate them into text-to-point-cloud tasks. As the pioneering work, Diffusion-PC~\cite{db1} introduces a novel probabilistic framework for generating 3D point clouds. By modeling the reverse diffusion process as a Markov chain, inspired by non-equilibrium thermodynamic principles, it effectively transforms noise distributions into structured point clouds. Some methods~\cite{db2,db3,db4} follow a "text-to-image-to-point-cloud" paradigm, wherein a 2D image is first generated through the diffusion model and then converted into a 3D point cloud. For instance, in STPD~\cite{Wu_2023_ICCV}, a sketch and text guided diffusion model for colored point cloud generation is proposed, which first generates a 2D image and then converts it into a 3D point cloud. This approach leverages the strengths of diffusion models in image generation and then uses additional techniques to transform the 2D image into a 3D point cloud. Alternatively, some approaches~\cite{db1,db6} synthesize point clouds directly in 3D by sampling points and using a diffusion model. These pre-trained 3D generative models, rich in prior knowledge, can potentially serve as classifiers in open-world scenarios. In this work, we validate this using Shap-e~\cite{db3} and GaussianDreamer~\cite{db4}, chosen for their speed and performance. Specifically, Shap-e~\cite{db3} is known for its ability to directly generate point clouds based on given prompts, while GaussianDreamer~\cite{db4} first generates a coarse point cloud and then optimizes it using 3D Gaussian Splatting. Both models could be used to generate point clouds that can serve as inputs for classification tasks in open-world scenarios.

\subsection{3D Represent Learning}
Initial approaches to feature extraction from point clouds relied on geometric methods utilizing local information like normals and curvature~\cite{fe1}. PointNet~\cite{fe3} is the first to use neural networks for spatial feature extraction. Inspired by the Transformer's success in image and text domains~\cite{fe12}, many studies~\cite{fe13,fe15,fe16,fe17} adopt its framework for 3D point cloud models, leveraging its attention mechanism to excel in capturing both local and global features. For instance, Point Transformer~\cite{zhao2021point} is designed to take advantage of the local geometric relations between the center point and its neighbors.  Moreover, multi-view~\cite{bpnet,yang20232d}, voxel~\cite{voxnet,choy20194d,pbnet}, and octree~\cite{wang2017cnn,wang2023octformer} representations are also popular forms of point cloud representation. Addressing the open-pose challenge in open-world scenarios~\cite{open-pose}, we compare two models: TAP~\cite{fe11}, which lacks rotation invariance, and TET~\cite{fe10}, which tackles it. TAP~\cite{fe11} is a novel 3D-to-2D generative pre-training method for point cloud models. TAP generates 2D view images from 3D point clouds using a cross-attention mechanism, significantly improving performance on downstream tasks like classification and segmentation. Moreover, TET's~\cite{fe10} core idea involves a nonlinear TTlayer~($l_{TT}$), mapping point clouds from $\mathbb{R}^{N \times 3}$ to $\mathbb{R}^{N \times 4 \times K}$ via steerable spherical neurons, embedding data into conformal space to ensure rotational and permutation equivariance.

\section{Methodology}
\subsection{Problem Statement}
Given $N_c$ categories with the names set $\mathbf{CN}=\{CN_{1},CN_{2}, ... ,\\ CN_{N_c}\}$ in a dynamic environment, the 3D open-world classification task involves assigning the correct category label in $\mathbf{CN}$ to each instance within a sample set $\mathbf{TS}\in \mathbb{R}^{N_t \times N_p \times3}$. 
Here, $N_t$ and $N_p$ refers to the number of point cloud test samples and the number of points in each sample. Typically, in open-world settings, real samples for these categories are unavailable for classifier training. To bridge this gap, external knowledge can be leveraged to support adaptation to these novel categories.

\subsection{Overall Framework }
As illustrated in Fig.~\ref{fig:network}, our framework introduces a training-free adaptation paradigm for open-world 3D classification comprising two components: the construction of an open-world classifier and inferring test samples.
This paradigm uniquely addresses environmental dynamics through three key mechanisms: 1) \textbf{Knowledge-Aware Anchor Synthesis}: We first leverage large language models (e.g., ChatGPT) to obtain rich semantic descriptions for each novel category name, which are then transformed into diverse 3D anchor samples using pre-trained text-to-3D generative models. Crucially, this process preserves the full 3D spatial information that would be lost in projection-based approaches.
2) \textbf{Dynamic Environment Adaptation}: To robustly handle pose variations and environmental dynamics, we employ a carefully designed augmentation strategy that generates geometrically transformed variants of each anchor sample. These augmented samples are then processed through a frozen pre-trained 3D encoder to obtain rotation-invariant feature prototypes for each category.
3) \textbf{Efficient Open-World Inference}: For each test sample, we compute its feature representation and determine its category via cosine similarity to our prototypical anchors. This lightweight comparison mechanism eliminates the need for costly retraining while maintaining adaptation capability to novel environmental conditions. 

The entire pipeline operates without retraining, making it particularly suitable for dynamic environments where new categories and viewing conditions may emerge frequently.
As shown in Tab.~\ref{tab:open-pose model_comparison}, this approach achieves superior performance. Fig.~\ref{fig:anchor_number} demonstrates its high efficiency - this is achieved with only 7 samples per class, significantly fewer than required by retraining approaches.

\subsection{Construct the 3D Open-World Classifier}

Usually, the category name $CN_{i}$ consists of only one or two words, they fail to offer sufficient guidance for generative models. 
Therefore, we propose employing LLMs to produce more detailed and informative descriptions $\mathbf{CD}$ of these category names. The procedure can be concisely summarized by the following formula:
\begin{equation}
\mathbf{CD}=\mathbf{LLMs}\left(Q, \mathbf{CN}\right), 
\end{equation}
where $Q$ is the prompt for LLMs, e.g. 

``\texttt{What does the shape of an} [$CN_i$] \texttt{look like?}''  
Specifically, when $CN_i$ is ``\texttt{Teddy Bear}'', $Q_i$ would be ``\texttt{What does the shape of a teddy bear look like?}'', and $CD_i$ would be ``\texttt{A teddy bear typically features a round head, large eyes, a small nose, and chubby body and limbs.}''.

\noindent\textbf{Anchor Sample Generation.} The original or detailed textual descriptions, denoted as $\mathbf{CN}/\mathbf{CD}$, are input into text-to-3D generation models $\mathbf{GMs}$ to generate anchor samples $AS$ for each category. In the following text, all instances of $\mathbf{CD}$ can be replaced with $\mathbf{CN}$. 
In order to improve the robustness of the classifier we aim to construct, we generate $N_a$ anchor samples for each $CD_i$ by modulating the $noise_j$ within the pre-trained generative model. The process can be succinctly described by the following formula:
\begin{equation}
AS_i^j=\mathbf{GMs}\left(noise_j, CD_i\right), \quad AS_i^j \in \mathbb{R}^{N_p^{'} \times 3},
\end{equation}
where $CD_i$ stands for the $i_{th}, i\in(0, N_c]$ category text description, $AS_i^j$ serves as the $j_{th}, j\in(0, N_a]$ anchor sample for the $i_{th}$ category, and $noise_j$ is the $j_{th}$ diffusion noise of text-to-3D generation models $\mathbf{GMs}$. In addition, $N_p^{'}$ donates the number of points in the anchor sample.

Considering that there may be geometric representation differences between anchor samples and test samples, such as size and number of points included, we employ three standard data augmentation techniques: farthest point sampling (FPS), origin shifting, and scaling and rotation, to minimize these discrepancies. The process can be formally expressed as follows:
\begin{equation}
\hat{AS_i^j} = \frac{\mathbf{FPS}(AS_i^j, \hat{N_p}) - \mathbf{Avg}(AS_i^j)}{\mathbf{L}(AS_i^j)}  \cdot R, \quad R \in \mathbb{R}^{3 \times 3},
\end{equation}
where $\mathbf{FPS}(\cdot, \hat{N_p})$ serves as the farthest point sampling operation, and $\hat{N_p}$ stands for the final number of sampling points. Additionally, $\mathbf{Avg}(\cdot)$ donates the operation to average coordinates of the point cloud, and $\mathbf{L}(\cdot)$ is to calculate the maximum Euclidean distance from any point in the point cloud to the centroid. Moreover, $R$ is the random rotation matrix.

\noindent\textbf{Anchor Sample Feature Distribution.} Different 3D representation models $\mathbf{RMs}$ have a significant impact on 3D open-world classifiers. Existing models are often limited to aligned datasets and overlook the influence of open-pose variations. After thorough exploration, we select TET~\cite{fe10}, a 3D representation model that focuses on the study of rotational variance, as the feature extractor for constructing our open-world classifier. To this end, the anchor sample feature $AF$ could be attained by the following formula:
\begin{equation}
AF_i^j=\mathbf{RMs}\left(\hat{AS_i^j}\right), \quad AF_i^j \in \mathbb{R}^{\hat{N_p} \times D},
\end{equation}
where $AF_i^j$ serves as the $j_{th}, j\in(0, N_a]$ anchor sample feature for the $i_{th}$ category, and $D$ donates the feature dimension. As shown in Fig.~\ref{fig:network}, the anchor sample feature distribution implies the classification ability of our 3D open-world classifier. The hyperplane separating the features acts as the decision boundary for our classifier. On the other hand, the region where the features of a test sample fall determines the predicted category of that test sample. In the following section, we offer a comprehensive explanation of how our constructed open-world classifier is employed to classify objects.

\subsection{Inference Phase}
Given the test sample set $\mathbf{TS}$, we apply the same data augmentation operations as those used for the anchor samples to minimize the discrepancy in geometric representations. The process can be succinctly described by the following formula:
\begin{equation}
\hat{TS_m} = \frac{\mathbf{FPS}(TS_m, \hat{N_p}) - \mathbf{Avg}(TS_m)}{\mathbf{L}(TS_m)}  \cdot R, \quad R \in \mathbb{R}^{3 \times 3},
\end{equation}
where $\hat{TS_m}$ refers to the $m_{th}, m\in(0, N_t]$ augmented sample. Accordingly, we input this feature into representation models $\mathbf{RMs}$ to obtain the corresponding test sample feature $TF_m$. The process can be formally expressed as follows:
\begin{equation}
TF_m=\mathbf{RMs}\left(\hat{TS_m}\right), \quad TF_m \in \mathbb{R}^{\hat{N_p} \times D}.
\end{equation}
In order to measure the cosine similarity $d_i^j$ between the feature representation of the current test sample $TS_m$ and that of each anchor sample $AF_i^j$, we apply the following formula:
\begin{equation}
d_i^j = 1 -  \frac{AF_i^j \cdot TF_m}{\|AF_i^j \|_2 \|TF_m\|_2},
\end{equation}
where $\|\cdot \|_2$ signifies the norm of the feature vector. Furthermore, the category name $CN_k$ associated with the anchor sample feature $AF_k^j$ that is most similar to the test sample feature $TF_m$ will be adopted as the predicted class $P_m$ for the test sample $TS_m$. This process is represented as follows:
\begin{equation}
P_m=CN_k, k=\arg\min_{i}(\{d_1^1, d_1^2, ..d_i^j\}_{i,j=1,1}^{i,j=N_c, N_a}),
\end{equation}
where $\arg\min_{i}(\cdot)$ denotes the index $i$ when the minimum value is taken.

\section{Experiment}
\subsection{Experiment Setup}
\subsubsection{Datasets.}
The effectiveness of the proposed method is validated on the public datasets ModelNet40$^{\ddagger}$, ModelNet10$^{\ddagger}$, and McGill$^{\ddagger}$,  proposed in OP3D~\cite{open-pose},  to assess performance under diverse geometric complexity.
The ScanObjectNN~\cite{uy-scanobjectnn-iccv19} dataset, featuring real-world scanned data with occlusions and noise across 15 categories, is used to evaluate the effect of pretrained representation models.

ModelNet40$^{\ddagger}$/10$^{\ddagger}$ 
evolves from standard ModelNet40/10~\cite{ds1} through random rotation of the samples, resulting in open-pose variations. ModelNet40 serves as a benchmark dataset within the domain of 3D point cloud classification. It originates from computer-aided design models and comprises high-quality 3D synthetic objects, which are devoid of noise and exhibit regular structural features. The dataset encompasses 40 categories of common objects, including pianos, tables, and airplanes, and contains a total of 12,311 3D objects, each represented by 10,000 points. These points are characterized by 3D coordinates (x, y, z) and may include additional features, such as normal vectors or color information (R, G, B). ModelNet10 is a specifically curated subset of ModelNet40, focusing on 10 common object categories, such as beds, monitors, and chairs, containing a total of 4,899 3D objects.

McGill$^{\ddagger}$~\cite{open-pose}, an enhanced variant of the McGill dataset~\cite{ds2}, created through random sample rotation and the removal of five categories that overlap with ModelNet40. This modified dataset contains 14 distinct categories of complex geometric shapes (e.g., ants, snakes, and hands) characterized by significant morphological diversity. Natural object categories pose challenges for current methods due to complex topological variations.

McGill$^{\ddagger}$~\cite{open-pose} is derived from McGill~\cite{ds2} by randomly rotating the samples in McGill and removing five categories that overlap with ModelNet40. McGill$^{\ddagger}$ includes 14 categories of complex geometric shapes, such as ants, snakes, and hands.
The objects in this dataset exhibit higher morphological diversity, particularly posing greater challenges when handling natural object categories.
Each sample consists of point coordinates (x, y, z) and topological information. 

\begin{table}[t]
\centering
\renewcommand{\arraystretch}{1.2} % Adjust row height
\setlength{\tabcolsep}{5pt} % Adjust column spacing
\caption{Comparison to SOTAs. Proj. indicates whether it is necessary to project the 3D samples into 2D images. $^{*}$ denotes that there is an overlap between test categories and training categories of the pre-trained representation model utilized.}
\begin{tabular}{lccccccc}
\toprule
           Methods          & Proj.         & \multicolumn{2}{c}{ModelNet10$^{\ddagger}$} & \multicolumn{2}{c}{McGill$^{\ddagger}$} \\ 
\cmidrule(lr){3-4} \cmidrule(lr){5-6} 
                               &                        &                        oAcc$\uparrow$  & mAcc$\uparrow$  & oAcc$\uparrow$  & mAcc$\uparrow$ \\ 
\midrule
    PointCLIP              & Y              & 17.7          & 16.4          & 12.2          & 13.3          \\ 
    ULIP & N & 14.4$^{*}$&13.8$^{*}$&14.8&16.1\\
    ReconCLIP  & Y   & 14.4$^{*}$ & 13.8$^{*}$     & 14.8     & 16.1          \\ 
    CLIP2Point  & Y    & 15.6$^{*}$  & 14.3$^{*}$     & 15.7     & 17.3          \\ 
    PointCLIPv2   & Y    & 19.9    & 18.2    & 27.8   & 28.9          \\ 
     OP3D-CLIP    & Y    & \underline{26.3}  & \underline{24.2}  & 31.3  & 34.6  \\ 
     OP3D-Diffusion    &Y   & 22.6  &21.7 & \underline{39.1}  & \underline{44.7}  \\

Ours                   & N          & \textbf{59.0} & \textbf{60.3} & \textbf{47.8} & \textbf{47.5} \\
\bottomrule
\end{tabular}
\label{tab:open-pose model_comparison}
\end{table}

\subsubsection{Evaluation Metric.}
Following SOTAs~\cite{zhang2023clip,huang2023clip2point,open-pose}, we evaluate the open-world classification performance using top-1 accuracy (oAcc) and mean accuracy (mAcc).
oAcc measures the proportion of samples for which the model's top prediction matches the ground truth, while mAcc calculates the average accuracy across all categories to assess the balance of performance among classes.

\subsubsection{Implement Details.}
All feature extraction and inference tasks are performed on an NVIDIA RTX 3090 GPU. % with CUDA 11.8.
For the point cloud diffusion model, we utilize a random seed \( s \sim \mathcal{U}(0, 50) \) to introduce diversity into the generated anchor point clouds of any category.
During the data augmentation stage, all the anchor samples and test samples are downsampled to 1024 points by the farthest point sampling.
For the feature output dimensions of different models, the TET pre-trained model outputs features with dimensions of (256, 1), and the TAP pre-trained model outputs features with dimensions of (256, 384).
To facilitate subsequent feature comparison and processing, 
max pooling along TAP's output second dimension unifies features to (256, 1).

\begin{figure}[t]  % [h] 表示尽可能在当前位置插入图片
  \centering
  \includegraphics[width=0.475\textwidth]{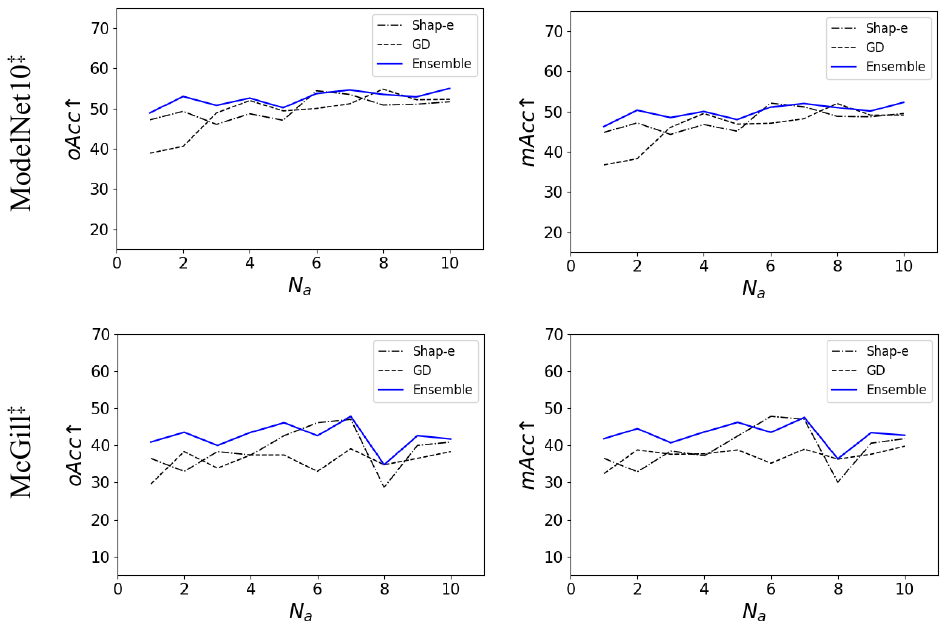}  % 插入图片并设置宽度为文本的80%
  \caption{Comparison on the Number of Anchor Samples $N_a$.}  % 图片的标题
  \label{fig:anchor_number}  % 设置图片标签，方便引用
\end{figure}

\subsection{Performance Comparison}
\label{chap:performance_comparison}

Our approach is compared  against six SOTA methods: PointCLIP~\cite{zhang2023clip}, ULIP~\cite{xue2023ulip}, ReconCLIP~\cite{qi2023contrast}, CLIP2Point~\cite{huang2023clip2point}, PointCLIPv2~\cite{Zhu2022PointCLIPV2}, and OP3D~\cite{open-pose}, on the open-pose classification datasets ModelNet10$^{\ddagger}$ and McGill$^{\ddagger}$.
To ensure fairness, we utilize the pre-trained models from the official GitHub repositories provided by the respective papers (TAP~\cite{fe11} and TET~\cite{fe10}) and strictly followed the original experimental settings.
For the tests on the McGill dataset, we employ a representation model trained on ModelNet40, as there is no category overlap between these datasets.
However, due to the overlapping categories between ModelNet40 and ModelNet10, we manually exclude the overlapping categories from ModelNet40 to form ModelNet30, and subsequently trained a new representation model on this adjusted dataset for the ModelNet10 tests.
The results presented in Section~\ref{chap:performance_comparison} and Section~\ref{chap:result_analysis} do not include the samples generated from LLM-generated prompts.
That is, all anchor samples are directly generated using the category names.
The impact of LLM-generated prompts on the results will be discussed in subsequent sections.

\textbf{Our approach exhibits notable enhancements in classification accuracy over current models in open-pose recognition tasks.}
The performance of our model on McGill$^{\ddagger}$  dataset and ModelNet10$^{\ddagger}$ dataset are shown in Table~\ref{tab:open-pose model_comparison}.
Notably, we achieved a \textbf{32.7\% improvement in oAcc} and a \textbf{36.1\% improvement in mAcc} on ModelNet10$^{\ddagger}$ dataset.
On McGill$^{\ddagger}$ dataset, our model also demonstrates competitive performance.
Compared to the current leading method, our approach \textbf{improve oAcc by 8.7\%} and \textbf{mAcc by 2.8\%}. 

\begin{table*}[t!]
\centering
\caption{Comparison of classification performance using anchor samples generated by different generative models on randomly rotated dataset ModelNet10$^{\ddagger}$.}
\begin{tabular}{ccccccccccc|cc}
\toprule
 Generative Model        & Bathtub       & Bed           & Chair         & Desk          & Dresser       & Monitor       & Night Stand   & Sofa          & Table         & toilet        & oAcc$\uparrow$           & mAcc$\uparrow$          \\ \hline
Shap-e   & \textbf{80.0} & \textbf{40.0} & 37.0          & 73.3          & 73.3          & 36.0          & \textbf{44.2} & 49.0          & \textbf{79.0} & \textbf{80.0} & 57.8          & 59.2          \\
GD       & 74.0          & 34.0          & \textbf{47.0} & 75.6          & \textbf{76.7} & \textbf{38.0} & 23.3          & 58.0          & 66.0          & 79.0          & 56.2          & 57.2          \\
Ensemble & \textbf{80.0} & 27.0          & 43.0          & \textbf{77.9} & 74.4          & 39.0          & 38.4          & \textbf{67.0} & 77.0          & 79.0          & \textbf{59.0} & \textbf{60.3} \\ 
\bottomrule
\end{tabular}
\label{tab:modelnet10 shap-e and GD results}
\end{table*}

\begin{table*}[t!]
\centering
\caption{Comparison of classification performance using anchor samples generated by different generative models on randomly rotated dataset McGill$^{\ddagger}$. Abbreviations used: Din. = Dinosaur, Dol. = Dolphin, Oct. = Octopus, Qua. = Quadruple, Spe. = Spectacle, Spi. = Spider.}
\resizebox{\textwidth}{!}{
\begin{tabular}{ccccccccccccccc|cc}
\toprule
 $\mathbf{RMs}$ & Ant & Bird & Crab & Din. & Dol. & Fish & Hand & Oct. & Pliers & Qua. & Snake & Spe. & Spi. & Teddy & oAcc$\uparrow$  & mAcc$\uparrow$  \\ \hline
Shap-e & 0.0 & \textbf{42.9} & \textbf{20.0} & 42.9 & 25.0 & 12.5 & \textbf{85.7} & \textbf{25.0} & 71.4 & 81.8 & 22.2 & \textbf{55.6} & \textbf{72.7} & 100.0 & 47.0 & 47.0 \\
GD & \textbf{20.0} & 28.6 & 0.0 & \textbf{57.1} & 25.0 & \textbf{25.0} & 14.3 & 12.5 & 71.4 & \textbf{90.9} & 22.2 & 33.3 & 45.5 & 100.0 & 39.1 & 39.0  \\ 
Ensemble & 0.0 & \textbf{42.9} & \textbf{20.0} & \textbf{57.1} & 25.0 & \textbf{25.0} & 57.1 & \textbf{25.0} & 71.4 & \textbf{90.9} & \textbf{33.3} & 44.4 & \textbf{72.7} & 100.0 & \textbf{47.8} & \textbf{47.5}  \\ 
\bottomrule
\end{tabular}
}
\label{tab:Mcgill shap-e and GD results}
\end{table*}

\begin{table*}[t!]
\centering
\caption{Comparative evaluation of classification performance under aligned-pose (ModelNet10) and open-pose (ModelNet10$^{\ddagger}$) settings across various representation models, with Shap-e used as the generative model. RMs denote the representation models used for extracting features.}
\resizebox{\textwidth}{!}{
\begin{tabular}{cccccccccccc|cc}
\toprule
                                         & $\mathbf{RMs}$ & Bathtub       & Bed           & Chair         & Desk          & Dresser       & Monitor       & Night Stand   & Sofa          & Table         & Toilet        & oAcc $\uparrow$ & mAcc$\uparrow$ \\ \hline
\multirow{2}{*}{ModelNet10$^{\ddagger}$} & TAP            & 24.0          & 11.0          & 16.0          & 5.8           & 9.3           & 23.0          & \textbf{65.1} & 36.0          & 3.0           & 25.0          & 21.5            & 21.8           \\
                                         & TET            & \textbf{80.0} & \textbf{40.0} & \textbf{37.0} & \textbf{73.3} & \textbf{73.3} & \textbf{36.0} & 44.2          & \textbf{49.0} & \textbf{79.0} & \textbf{80.0} & \textbf{57.8}            & \textbf{59.2}           \\ \hline
\multirow{2}{*}{ModelNet10}              & TAP            & 24.0          & \textbf{55.0} & \textbf{63.0} & 10.5          & 51.2          & \textbf{78.0} & \textbf{83.7} & \textbf{79.0} & 31.0          & 78.0          & 57.4            & 55.3           \\
                                         & TET            & \textbf{74.0} & 47.0          & 47.0          & \textbf{74.4} & \textbf{68.6} & 44.0          & 54.7          & 34.0          & \textbf{73.0} & \textbf{85.0} & \textbf{59.1}   & \textbf{60.2}                        \\ 
\bottomrule
\end{tabular}
}
\label{tab:TAP nad TET on open-pose and aligned-pose}
\end{table*}

\begin{figure}[t]
    \centering
    \begin{subfigure}[b]{0.22\textwidth}
        \includegraphics[width=\textwidth]{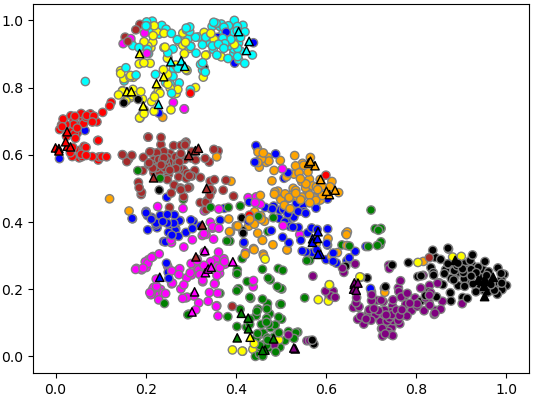}
        \caption{TET on aligned-pose}
    \end{subfigure}
    \hspace{0.01\textwidth}
    \begin{subfigure}[b]{0.22\textwidth}
        \includegraphics[width=\textwidth]{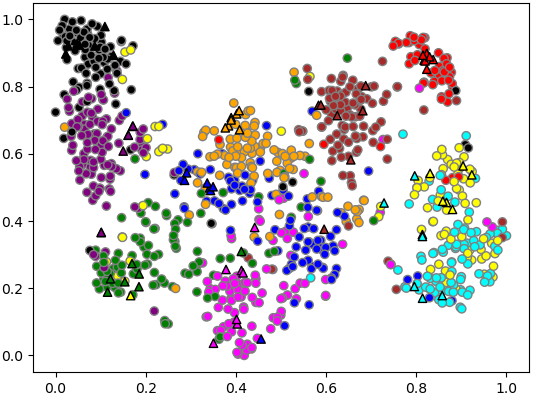}
        \caption{TET on open-pose}
    \end{subfigure}
    \vspace{0.01\textwidth}
    
    \begin{subfigure}[b]{0.22\textwidth}
        \includegraphics[width=\textwidth]{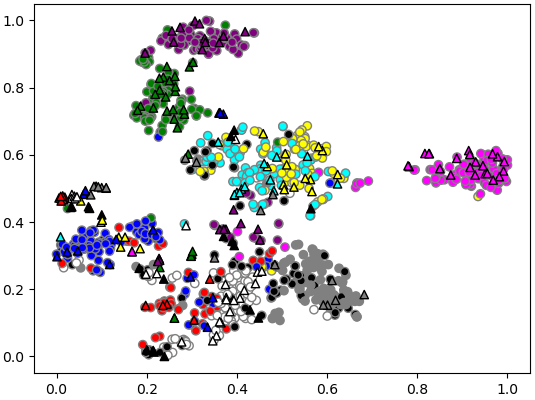}
        \caption{TAP on aligned-pose}
    \end{subfigure}
    \hspace{0.01\textwidth}
    \begin{subfigure}[b]{0.22\textwidth}
        \includegraphics[width=\textwidth]{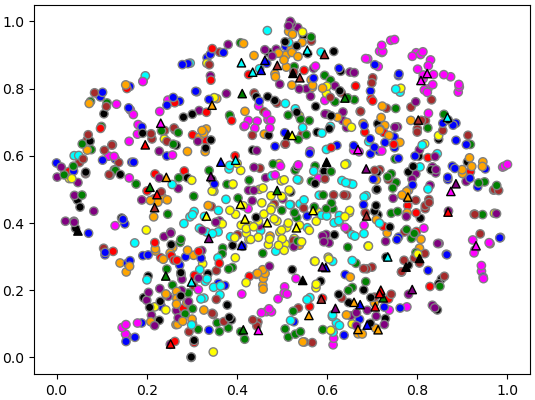}
        \caption{TAP on open-pose}
    \end{subfigure}

    \caption{t-SNE visualizations of feature representations extracted by TAP and TET under aligned-pose and open-pose settings on the ModelNet10 dataset. Different colors indicate different object categories according to the ground truth labels. Triangles represent generated anchor samples, while circles denote original dataset samples.}
    \label{fig:TAP_TET_TSNE}
\end{figure}

\begin{table*}[t]
\centering
\renewcommand{\arraystretch}{1.2} 

\caption{Impact of TET pre-trained on different datasets when evaluated on ModelNet10$^{\ddagger}$}
\begin{tabular}{lccccc}
\toprule
Pre-training Source      & Training sample numbers & Categories numbers & Overlapping Categories numbers  &    oAcc$\uparrow$       &    mAcc$\uparrow$ \\
\midrule
ModelNet40     & 9840 & 40&       10             &        90.2       &     90.3   \\
ModelNet30    &  6049 & 30&       0          &         59.0            &     60.3         \\
ScanObjectNN   & 2309   & 15&        7        &        54.6       &     52.0   \\
\bottomrule
\end{tabular}
\label{tab:different_pretrain_comparison}
\end{table*}

\subsection{Analysis of Results}
\label{chap:result_analysis}

The performance of the proposed method is influenced by several factors, including the number of anchor points, the selection of generative models, and 3D representation models. 
This section presents a quantitative analysis of each factor's impact, showcasing our approach's effectiveness and robustness. 

\subsubsection{Impact of Anchor Sample Size $N_a$}

We examine the impact of varying the number of anchors generated while fixing the dataset, representation backbone (TET), and generation model to evaluate its effect on classification performance.
Three distinct anchor generation methods are evaluated: Shap-e, GaussianDreamer (GD), and their mixed (Ensemble).
\textbf{As shown in Fig.~\ref{fig:anchor_number}, classification accuracy exhibits positive correlation with $N_a$, eventually reaching stable.} However, on McGill$^{\ddagger}$, performance slightly decreased at 8 anchor samples, likely attributable to instability when anchor counts are low—where individual sample variance disproportionately influences feature statistics. This effect diminishes as $N_a$ increases. 
While more anchor samples generally improve performance, the marginal gain decreases beyond a point. This must be balanced against linearly increasing computational overhead during generation. 
Based on empirical convergence and computational trade-offs,  we set $N_a=7$ as the optimal anchor points across all the experiments. 

\subsubsection{Comparison on Generative Models}

The experiment maintains the constant dataset, representation model (TET), and number of anchor points, altering only the generative models to assess their impact on classification performance.
Two generative models are selected: Shap-e and GD.
Both models can generate point clouds based on the given prompts, but they differ in methodology.
Shap-e directly generates a point cloud, whereas GD first generates a coarse point cloud and then optimizes it using 3D Gaussian Splatting.
In terms of generation time, Shap-e requires 7 seconds per sample, while GD take 15 minutes per sample. The results on ModelNet10$^{\ddagger}$ dataset and McGill$^{\ddagger}$ dataset are shown in Table~\ref{tab:modelnet10 shap-e and GD results} and Table~\ref{tab:Mcgill shap-e and GD results}, respectively.
Overall, Shap-e outperforms GD in terms of classification accuracy on both datasets, but it does not consistently lead in every category.
Both generative models have their strengths. On ModelNet10$^{\ddagger}$, Shap-e excels in some categories, e.g. \textit{bathtub}, \textit{bed}, and \textit{night stand}, while GD performs better in others, e.g. \textit{chair}, \textit{dresser}, and \textit{sofa}.
On McGill$^{\ddagger}$, Shap-e performs significantly better in \textit{bird}, \textit{hand}, \textit{spectacle}, and \textit{spider}, while GD outperforms Shap-e in \textit{ant}, \textit{dolphin}, \textit{fish}, and \textit{quadruped}.
We further conduct experiments by combining anchor samples generated by Shap-e and GD, referred to as Ensemble. 
The results demonstrate that the integration of anchor samples generated by both generative models results in enhanced overall performance. Significantly, the category-specific accuracy obtained by the Ensemble closely aligns with the leading generative model for each category. These results imply that \textbf{the diversity inherent in samples produced by 3D generative models plays a beneficial role in augmenting the efficacy of the classification task}.

\begin{table*}[t]
\centering

\caption{Comparison of prompts and the classification accuracies of samples generated for `dresser'.}
\begin{tabular}{lc}
\toprule
Category Text Description $\mathbf{CD}$  & oAcc of `Dresser'$\uparrow$ \\ \hline
``\texttt{A dresser.}'' & 73.3 \\ %10.5
``\texttt{A dresser with five drawers and circular handles.}'' & 68.6 \\ 
``\texttt{An open design dresser with five exposed drawers.}'' & 72.1 \\ 
``\texttt{A small bedside dresser with two drawers}''. & 68.6 \\ 
``\texttt{A vintage dresser with carved details and two doors.}'' & 76.7 \\ 
``\texttt{A decorative dresser with floral carvings and four drawers.}'' & 57.0 \\ 
``\texttt{A wooden dresser with a built-in mirror and six drawers.}'' &\textbf{86.0} \\ 

\bottomrule
\end{tabular}

\label{tab:prompts_acc}
\end{table*}

\begin{table*}[t]
\centering
\caption{Comparison of classification result on ModelNet10$^{\ddagger}$  using Shap-e with Category Name ($\mathbf{CN}$) and ChatGPT ($\mathbf{CD}$) prompts.}

\resizebox{0.9\textwidth}{!}{
\begin{tabular}{ccccccccccc|cc}
\toprule
   & Bathtub & Bed           & Chair         & Desk          & Dresser       & Monitor       & Night Stand   & Sofa          & Table         & Toilet        & oAcc$\uparrow$           & mAcc $\uparrow$          \\ \hline

$\mathbf{CN}$ & 80.0    & 40.0          & \textbf{37.0} & \textbf{73.3} & 73.3          & 36.0          & \textbf{44.2} & \textbf{49.0} & 79.0          & \textbf{80.0} & 57.8          & 59.2          \\
$\mathbf{CD}$    & 82.0    & \textbf{44.0} & 50.0          & 74.4          & \textbf{72.1} & \textbf{31.0} & 43.0          & 43.0          & \textbf{82.0} & 86.0          & \textbf{59.5} & \textbf{60.8} \\
\bottomrule
\end{tabular}
}
\label{tab:prompt_generation_class}
\end{table*}

\subsubsection{Comparison on Representation Models} 
We compare two 3D representation models in this experiment: TAP, which lacks rotation invariance, and TET, which is designed to address this issue.
The comparison is conducted on the aligned-pose dataset ModelNet10 and the open-pose dataset ModelNet10$^{\ddagger}$.
In addition, we only utilize Shap-e as the generative model.
Experimental results, shown in Table~\ref{tab:TAP nad TET on open-pose and aligned-pose}, demonstrate that on the aligned-pose dataset, TAP and TET exhibit comparable performance.
This implies that the two models, despite being designed with distinct approaches, display minimal performance disparities on the aligned-pose dataset.
Notably, on the open-pose dataset, TET significantly outperforms TAP across the majority of categories.
For instance, TET shows a significant improvement over TAP’s oAcc of 36.3\% and mAcc of 37.4\%.

It can be observed that \textbf{designing specific structures to address point cloud rotation invariance effectively enhances the model's robustness against rotation}.
The results show that TET, which incorporates such structures, is unaffected by point cloud rotations, as evidenced by the minimal difference in performance between the aligned-pose and open-pose datasets.
In fact, TET exhibits a slight performance drop on the open-pose dataset, with oAcc and mAcc decreasing by 1.3\% and 1.0\%, respectively. 
In contrast, TAP, which uses a Transformer-based framework without specialized structures for rotation invariance, achieves results comparable to those of TET on the aligned-pose dataset.
However, its performance drops sharply on the open-pose dataset, underscoring the need for rotation-invariant design to maintain robustness.

To provide a more intuitive understanding of this issue, as shown in Fig.~\ref{fig:TAP_TET_TSNE}, we visualize the feature representations of TAP and TET on the ModelNet10 dataset using t-SNE under both aligned-pose and open-pose settings.
In the aligned-pose scenario, TAP and TET exhibit similarly well-clustered feature distributions.
The generated samples (marked with triangles) also contribute effectively to category separation.
However, under the open-pose setting, TAP’s t-SNE plot becomes cluttered and disorganized, indicating that its features are less distinguishable due to sensitivity to rotation.
In contrast, TET maintains clear and compact clustering, further confirming its robustness to arbitrary rotations.

\subsubsection{Effect of  Representation Models on Pre-trained Datasets} 
We study the impact of pre-training representation models for open-world classification by assessing the TET model, pre-trained on ModelNet40, ModelNet30, and ScanObjectNN, on the ModelNet10$^{\ddagger}$ dataset, as illustrated in Table~\ref{tab:different_pretrain_comparison}.
When the pre-training dataset fully overlaps with the test categories (i.e., ModelNet40 includes all ModelNet10 categories), the model achieves an oAcc of 90.2\%, a result competitive with those obtained via fully supervised learning.
This demonstrates that our generated samples contain valid and discriminative category-level information, and also confirms the effectiveness of the proposed pipeline.
However, when forming ModelNet30 by removing overlapping categories,  performance drops significantly, indicating the challenge category-level openness poses, even for rotation-invariant models.
Furthermore, using ScanObjectNN, with fewer categories and samples and a distinct data distribution (real-world scans instead of CAD models), the model still achieves an oAcc of 54.6\% despite 7 overlapping categories, indicating that out-of-distribution (OOD)  issues persist in rotation-invariant models.
On the other hand, the result highlights that \textbf{high category overlap alone does not guarantee high performance when the pre-training and testing distributions differ significantly}.

\begin{figure}[t!]  % [h] 表示尽可能在当前位置插入图片
  \centering
  \includegraphics[width=0.49\textwidth]{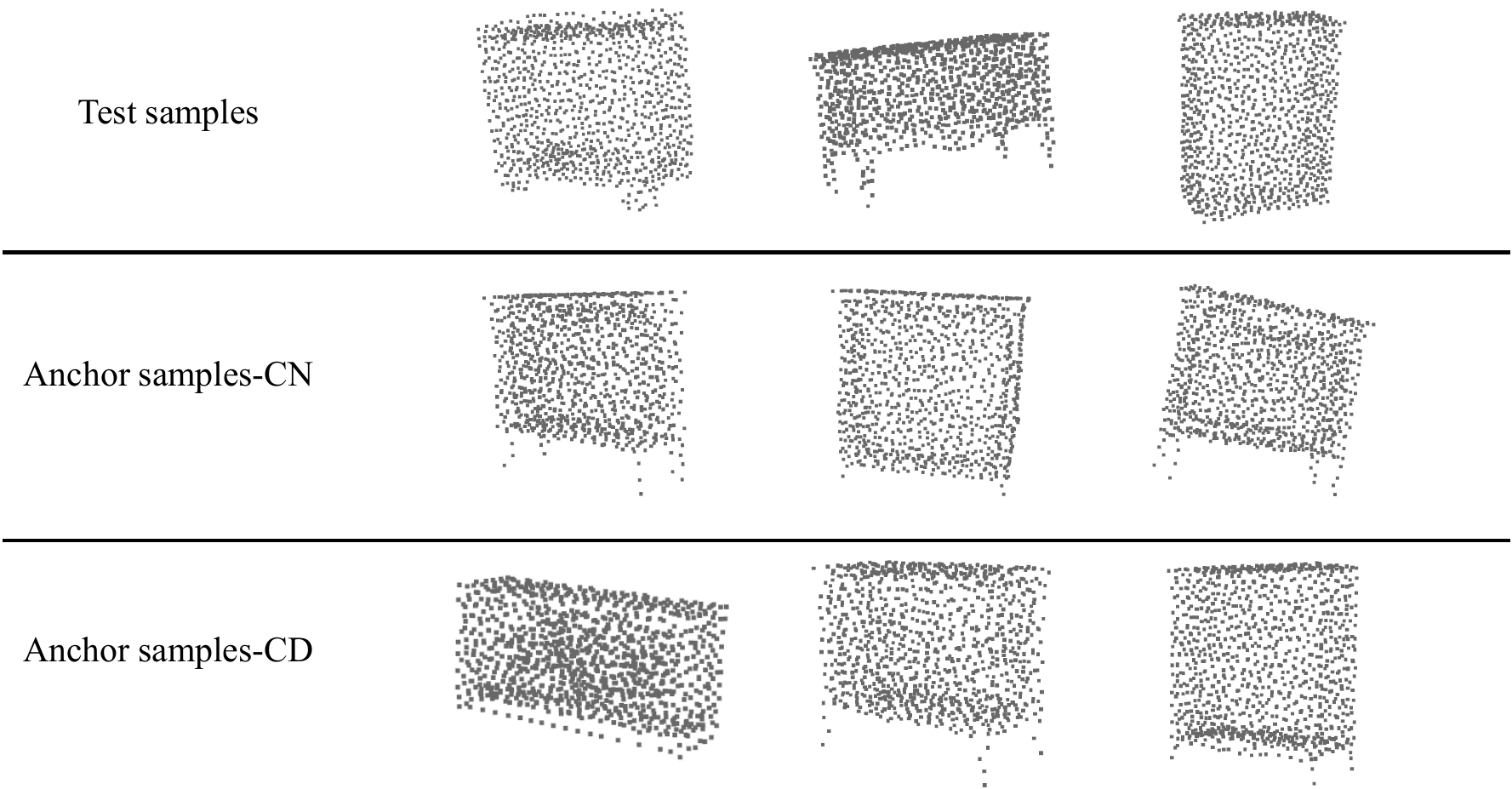}  % 插入图片并设置宽度为文本的80%
  \caption{Visualization comparison of "dresser" point clouds from original category names ($\mathbf{CN}$) and ChatGPT-generated prompts ($\mathbf{CD}$), which enhance anchors' diversity.}
  \label{fig:gpt_modelnet10}  % 设置图片标签，方便引用
\end{figure}

\subsection{Discussion on LLM-Generated Category Descriptions}

In our experiments, some categories have consistently low classification accuracy, likely due to a lack of sample diversity. We hypothesize that more varied samples could enhance results by optimizing descriptions for specific categories.
Taking the dresser category from the ModelNet10$^{\ddagger}$ dataset as an example, which has notably low accuracy, we employ ChatGPT to generate multiple descriptive phrases tailored to the dresser category.
These descriptions followed the format “a dresser that looks like ... .”
Using these optimized descriptions, we apply the Shap-e to create multiple new point clouds for the dresser category based on each description.
The experimental results are presented in Tables~\ref{tab:prompts_acc}, and the comparison between anchor points generated using ChatGPT-generated descriptions ($\mathbf{CD}$) versus category names ($\mathbf{CN}$) is shown in Fig.~\ref{tab:prompt_generation_class}.
The results demonstrate that appropriate descriptions significantly enhanced classification performance for the dresser category, while some inappropriate descriptions led to a decline in performance.
However, since we did not know beforehand which descriptions would yield the best results, in subsequent experiments, we generated anchor samples using each description for all categories to further improve performance.

We apply this approach on the ModelNet10$^{\ddagger}$ dataset, generating LLM-generated category descriptions to create anchor samples for classification tasks. 
From the results, shown in Table~\ref{tab:prompt_generation_class}, the oAcc and mAcc are increased by 1.7\% and 1.6\%, respectively. 
However, for many individual categories, we observe that the performance is worse compared to the results generated using only class names.
\textbf{The observed performance fluctuations are more likely due to differences in how well the prompts align with the core visual features of the target category.}
However, this also causes interference between categories, such as generating a dresser that closely resembles a table.
We exclude these results from method comparisons due to the instability of LLM-generated descriptions. 
Hence, improving control over LLM-generated descriptions for better results is a promising future direction.

\section{Conclusion}

This paper proposes a pipeline that leverages 3D point cloud generative models for open-world 3D classification. Unlike existing methods that rely on 2D priors and projection, our approach directly utilizes prior knowledge from 3D generative models, avoiding projection-induced instability in open-pose scenarios. The method is training-free, rotation-invariant, and capable of adapting to novel categories with only a few generated samples—an advantage especially valuable given the high cost of collecting real-world 3D data.

We further examine how anchor sample quantity, diversity, and feature space design affect performance, and explore the role of LLM-generated prompts. Performance variability across prompts is found to stem not from linguistic ambiguity, but from differences in how well prompts capture the visual and structural essence of the target category. Future work may explore improving prompt strategies to generate more diverse and representative samples that better reflect core category features.

\bibliographystyle{ACM-Reference-Format}
\bibliography{sample-base}

\appendix

\end{document}